# Phase Transitions in Knowledge Compilation: an Experimental Study

Jian Gao[1], Minghao Yin[1], and Ke Xu[2]

[1] College of Computer Science
Northeast Normal University
Changchun, 130024, China.
jiangao.cn@hotmail.com; ymh@nenu.edu.cn

[2] State Key Lab. of Software Development Environment
Beihang University
Beijing, 100191, China.
kexu@nlsde.buaa.edu.cn

**Abstract.** Phase transitions in many complex combinational problems have been widely studied in the past decade. In this paper, we investigate phase transitions in the knowledge compilation empirically, where DFA, OBDD and d-DNNF are chosen as the target languages to compile random $k$-SAT instances. We perform intensive experiments to analyze the sizes of compilation results and draw the following conclusions: there exists an easy-hard-easy pattern in compilations; the peak point of sizes in the pattern is only related to the ratio of the number of clauses to that of variables when $k$ is fixed, regardless of target languages; most sizes of compilation results increase exponentially with the number of variables growing, but there also exists a phase transition that separates a polynomial-increment region from the exponential-increment region; Moreover, we explain why the phase transition in compilations occurs by analyzing microstructures of DFAs, and conclude that a kind of solution interchangeability with more than 2 variables has a sharp transition near the peak point of the easy-hard-easy pattern, and thus it has a great impact on sizes of DFAs.

## Introduction

Phase transitions, as a kind of well-known phenomenon in artificial intelligence, have attracted a great mount of attention since the paper [1]. It is claimed that all NP-complete problems have a critical point that separates overconstrained and underconstrained regions, and soluble-to-insoluble phase transition occurs at this point. Successive studies have shown that phase transitions widely exist in complex combinational (optimization) problems [2-8]. Some representative problems are random $k$-SAT, whose phase transition point is measured by the ratio of the number of clauses to that of variables, and random Constraint Satisfaction Problems (CSPs), where in the RB model [2] an exact transition point was proved whose instances are hard to solve. Furthermore, recent investigations on more complex problems have demonstrated that there are phase transitions in QBF [5,6] and planning [7] which are PSPACE-complete, and even in problems that are EXPSPACE-complete [8]. In addition, some real-world problems, such as TSP [9] and manipulation problem [10], have been shown to have phase transition phenomena. In fact, it is worth mentioning that soluble-to-insoluble phase transition studied most widely is only one kind of transition. Many other properties of combinational problems suffer phase transitions. As an example Bailey et al. [11] have addressed phase transitions in #SAT, of which the decision problem is PP-complete. As second example phase transitions of backtrack-free instances have been studied [12].

The phase transitions mentioned above are always accompanied with the transitions of CPU runtimes. Namely, algorithms will suffer an easy-hard-easy pattern when solving those problems. Instances around the soluble-to-insoluble transition points are hard to solve by both systematic search algorithms and local search algorithms with other instances easy to solve. However, we will show that easy-hard-easy patterns are not only expressed in terms of the time, but also in terms of the space. That is phase transitions in knowledge compilation. Knowledge compilation [13] is used to compile all (or a part of) the solutions of a problem into a tractable language, and has been employed in many areas ranging from planning, diagnosis to formal verification and production configuration. In the past decades, many target languages of knowledge compilation have been proposed for compiling SAT instances and CSPs, such as Ordered Binary Decision Diagram (OBDD), prime implicates (PI), horn approximation, deterministic, Decomposable Negation Normal Form (d-DNNF), Deterministic Finite-state Automaton (DFA) and AND/OR MDD [14-18]. A knowledge compilation map [19] has been presented to analyze the succinctness of the target compilation languages, as well as identify tractable queries of them. Furthermore, easy-hard-easy patterns in target languages have also been shown in the early studies [20]. Schrag and Crawford studied phase transitions in compiling 3-SAT instances to PIs. They claimed that PIs of different lengths exhibit phase transitions that share many of the characteristics similar with satisfiability phase transition, and showed the critical point occurs when the ratio ($r$) of #clauses ($m$) to #variables ($n$) is around 2.0. While recent studies have proposed many more succinct languages [3], such as OBDD, d-DNNF and DFA. OBDD is a kind of binary decision diagram with a fixed variable ordering; d-DNNF is a subset of Negation Normal Form (NNF) that satisfies determinism and decomposability; DFA can be viewed as a special case of the OBDD, whose structures are more regular. Differ from PIs, these languages covert solutions into more compact forms using the property of solution symmetry. Moreover, they can compile large scale instances, and are more popular than other target languages in the knowledge compilation map mentioned above.

This paper focuses attention on the easy-hard-easy pattern with respect to space, and studies the easy-hard-easy pattern in knowledge compilation empirically, where OBDD, d-DNNF and DFA are employed as target languages. We concentrate on randomly generated $k$-SAT model, and discuss the phase transition phenomena that occur in compiling random SAT instances generated by various values of parameters into three target languages. We exhibit experimental results on compiling $k$-SAT instances into OBDD and d-DNNF respectively showing that these two languages suffer the easy-hard-easy pattern, where sizes of compilation results increase as the ratio of clauses to variables grows from 0 but decrease when the ratio exceeds a certain point. Hence, the phase transition occurs. We also observe a phase transition of polynomial-exponential increment of sizes when compiling instances with very small ratio of clauses to variables. Furthermore, we analyze the structural transition of DFAs empirically to show the reason why the easy-hard-easy pattern occurs. The experimental analysis shows that a sharp transition on the existence of a kind of paths, which we call multi-interchangeable paths, occurs around peak points of the easy-hard-easy pattern.

## Background of Random $k$-SAT Model

It is well-known that $k$-SAT is NP-complete for $k>2$. Many academic and real-world problems can be transformed into $k$-SAT and solved by the efficient SAT solvers. So designing search algorithms to solve SAT problems efficiently is an essential issue in AI. To evaluate SAT algorithms, Mitchell et al. [21] introduced a model for generating random SAT instances, which can produce hard instances as it was claimed. Compared with SAT instances from industrial area, this model can obtain large amount of instances whose size and difficulty can be controlled by parameters. As a result, it has become a canonical benchmark to test various SAT algorithms. Meanwhile, theoretical analyses on the random $k$-SAT model have been made [3], for example, the properties of phase transitions have been identified that are widely used to analyze average computational complexity.

In the model discussed in [21], there are 3 parameters $k$, $n$, $m$ that are employed to generate SAT instances, where $k$ is the length of clauses, $n$ is the number of variables, and $m$ is the number of clauses. We use a ratio $r$ to represent $m/n$. Instances are produced by generating $m$ clauses uniformly and independently, where each clause is generated by selecting $k$ variables without replacement from $n$ variables and negating each variable with probability 0.5. All clauses generated are of the same length.

Empirical and theoretical studies have demonstrated that there are easy-hard-easy patterns when solving the random $k$-SAT instances. Lots of experiments have shown that the soluble-to-insoluble phase transition will occur when $r$ is up to a fixed value, for example, $r$ is about 4.3 in 3-SAT [21], and mean runtimes for finding a solution by SAT algorithms are often extremely high for solving instances in the phase transition region. As $r$ increases from 0, mean runtimes increase first but decrease when $r$ is larger than the value at the transition point. However, it seems to be difficult to locate transition points of $k$-SAT by theoretical proof, as only lower and upper bounds have been found for the points [3]. Similar phenomenon also exists in #SAT [11], which is a counting problem. In the general case, #SAT is #P-complete, which is considered to be harder than NP-complete. Algorithms for solving this problem count the number of solutions in the given SAT instance. Mean runtimes of several #SAT solvers have been investigated, and the results show that those solvers also suffer the easy-hard-easy pattern, but the hardest regions of those #SAT algorithms are different. For instance, it was reported that an algorithm, called Counting Davis-Putnam [22], reaches its peak runtime when $r=1.2$ on the experiments of #3-SAT, while the peak point of another algorithm, called Decomposing Davis-Putnam [23], occurs when $r=1.5$.

In the following sections, we show that the knowledge complication also undergoes the easy-hard-easy pattern.

## Basic Conjecture

Let $P_k^{nr}$ be a random $k$-SAT instance generated with $n$ variables and $rn$ clauses, and let $X_L(P_k^{nr})$ be the size of the compilation result, where $L$ is the target language to which $P_k^{nr}$ is compiled (we regard the number of nodes as the size, when $L$ is OBDD or d-DNNF).

**Definition 1.** *Given n, r, k and a target language L, average size* $E(X_L(P_k^{nr}))$ *is defined as follows:*

$$E(X_L(P_k^{nr})) = (\sum X_L(P_k^{nr}))/|P_k^{nr}|$$

*where $P_k^{nr}$ is any problem in the set made up of all possible instances that can be generated by the random SAT model with parameters n, rn, k, and $|P_k^{nr}|$ is the number of all possible instances.*

Because the structures of the target languages are very complex, it seems hard to analyze average behaviors (i.e. $E(X_L(P_k^{nr}))$) of them theoretically. However, the following conjecture seems available, as it can be supported by a number of experiments.

**Conjecture 1.** *For every integer $k \geq 3$, given the number of variables n and a target language L that is a subset of DNNF, there exists a positive real number $r_c$ such that*
For each pair $(r_1,r_2)$, if $r_1<r_2<r_c$, then
$$E(X_L(P_k^{nr_1})) < E(X_L(P_k^{nr_2})).$$
For each pair $(r_1,r_2)$, if $r_c<r_1<r_2<r_p$, then
$$E(X_L(P_k^{nr_1})) > E(X_L(P_k^{nr_2})).$$
*where $r_p$ is the soluble-to-insoluble phase transition point of k-SAT.*

The conjecture describes that when $r$ ranges from 0 to $r_p$, the sizes of compilation results increase until $r$ reaches $r_c$ and then decrease.

## Experimental Results on Random *k*-SAT

In this section, we analyze phase transitions in random *k*-SAT compilation using OBDD and d-DNNF as target languages. The compilers we choose for the target languages are the state-of-the-art compilers, where d-DNNF compiler was made by Darwiche (It is available at: http://reasoning.cs.ucla.edu/c2d/) and the OBDD compiler based on CUDD package was developed by Narodytska and Walsh for their work of constraint and variable ordering heuristics for compiling configuration problems [14]. According to the knowledge compilation map, all the languages in the map are subsets of NNFs. Because NNFs can be described by directed acyclic graphs, we can represent OBDD and d-DNNF using directed acyclic graphs, of which the sizes are measured by the number of its nodes and edges. However, it can be observed that the number of nodes and the number of edges change in the same manner when we compile *k*-SAT instances ranging $r$ from 0 to $r_p$, and the peak points are also in the same position. Since we only care about the changing manner and the points of maximum sizes, with the fact discussed above, we count the number of nodes as the size of d-DNNF (OBDD), though the number of edges is usually used to evaluate the size of d-DNNF.

In the following experiments, we generate SAT instances using random *k*-SAT model and compile them into OBDDs and d-DNNFs respectively. For each combination of *k* and *n*, we vary *r* from 0 to the corresponding $r_p$ (relative to *k*). To obtain more exact results, we produce 1000 instances for each *r*, and calculate average nodes of the compilation results. Because of runtime limitations, we only take *k* ranging from 2 to 6 and *n* ranging from 20 to 60.

### Experiments on Random 3-SAT

The first set of experimental results concerns random 3-SAT instances. Fig. 1 depicts the easy-hard-easy pattern when compiling instances into d-DNNFs. We also locate the phase transition point in random 3-SAT compilations. There are 3 curves in Fig. 1, each of which corresponding to a different *n* ranging from 40 to 60. The peak points of those curves are with same value of the ratio *r*, which is 1.8. This point is far from the soluble-to-insoluble transition point where *r* is about 4.3. Fig. 2 describes the experimental results of OBDDs. Three groups of instances have been generated for the OBDD experiments too. But there is a small difference in the generation parameters with *n* varying from 20 to 40, since the OBDD compiler needs more time to finish the experiments. Note that the curves of OBDDs have the same behaviors as curves in Fig. 1. Especially, the peak sizes occur at the same point where *r* is 1.8. However, as we mentioned in pervious section, the critical point of PIs is slightly larger than 1.8, we surmise that because OBDDs and d-DNNFs utilizes the compactness of solutions, positions of critical points for these two languages differ from those for PIs.

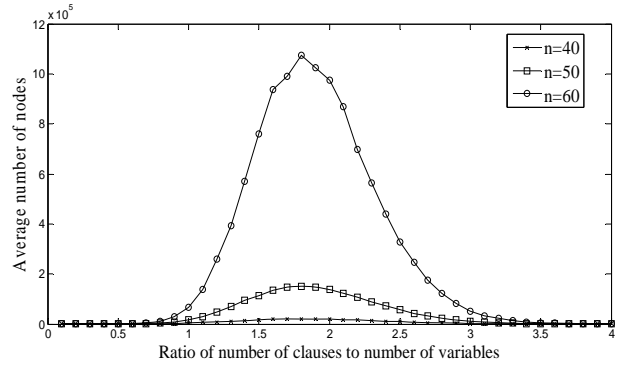

**Fig. 1.** Average # nodes of d-DNNFs, where 3-SAT instances were compiled.

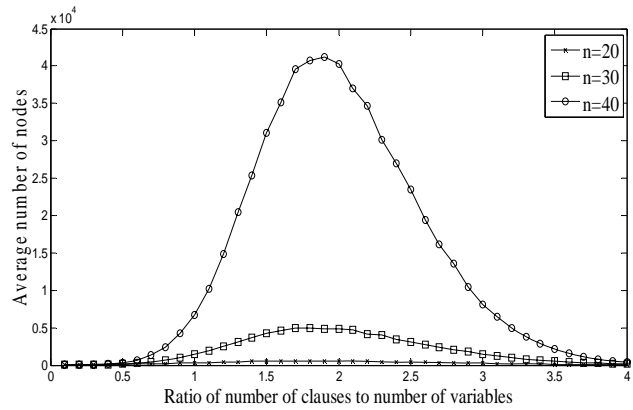

**Fig. 2.** Average # nodes of OBDDs, where 3-SAT instances were compiled

Interestingly, we exhibit that when $r$ ($r>0.3$ for 3-SAT) is fixed, the sizes of target languages increase exponentially as the number of variables $n$ grows linearly. We take 6 values of $r$ between 0 and $r_p$ uniformly. For each $r$, we generate 1000 3-SAT instances varying $n$ from 10 to 60 at increments of 5, and convert them into d-DNNFs. Compilation results are shown in Fig. 3. Curves with different values of $r$ are all nearly linear when the logarithmic vertical axis is used, so we believe that the size grows exponentially in the general cases. However, the slope of each curve depends on $r$. When $r$ is close to 1.8, the slope grows larger, and the curve of $r=1.8$ has the biggest slope among the 6 curves. As a result, though the average sizes increase exponentially, the sizes around phase transition regions grow fastest.

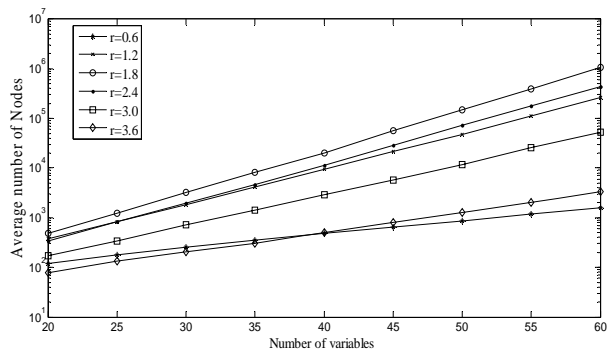

**Fig. 3.** Comparison of increments on the average #nodes of d-DNNFs with $n$ ranging from 20 to 60.

We also observe that there is another phase transition in compilation. That is, the sizes increase from polynomially to exponentially as *r* grows. Fig. 4 shows this transition, where double logarithmic axes are employed to identify the polynomial increment-to-exponential increment. We surmise that the sizes for 3-SAT increase polynomially when *r*<0.3 as the curves are nearly linear, while exponentially when *r*>0.3, and we believe that there is a phase transition separates the polynomial and exponential sizes. It is quite similar to the classical phase transition, where the phase transition in structures of solutions separates the tractable and intractable regions [12].

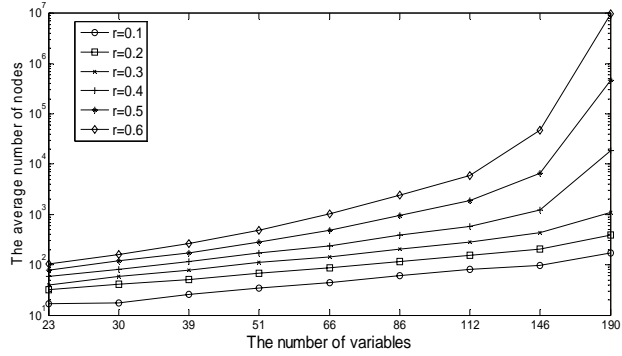

**Fig. 4.** Increments on the average number of nodes using d-DNNFs as the target language, where *r*=0.1 to 0.6 are tested as the number of variables grows exponentially ($n=1.3^{12},…,1.3^{20}$).

**Experiments on Random *k*-SAT**

We then show that phase transitions also exist in compiling *k*-SAT. Experimental results on *k*-SAT instances are depicted in Fig. 5. We generate instances by fixing *n* to 20, varying *k* from 2 to 6 and *r* from 0 to the corresponding $r_p$. Each curve in Fig. 5 (a) corresponds to a different value of *k*. We can also observe that the phenomenon of easy-hard-easy pattern exists in each curve, where the peak points are relative to *k*. Fig. 5 (b) exhibits the compilation results on OBDDs, where the same behaviors can be obtained as those on d-DNNFs, too. Besides, curves in the two figures demonstrate that $r_c/r_p$ becomes smaller as *k* increases.

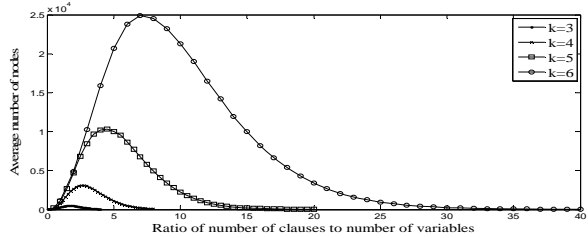

(a)

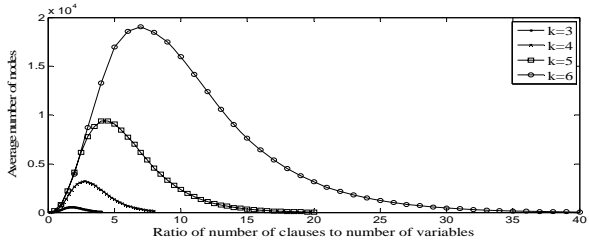

(b)

**Fig. 5.** Average #nodes of d-DNNFs (a) and OBDDs (b), where *k*-SAT instances were compiled with *k* ranging from 3 to 6.

We close this section by arguing that not only the size of compilation results undergoes the easy-hard-easy pattern, but also the runtime cost in compilations does, In [16], similar experimental results on d-DNNF for compiling 3-SAT instances have been depicted, where the peak point of runtimes is also fixed at $r$=1.8. As a result, it is hard to compile instances near the phase transition region but quite easier for other instances, including those hard to find a solution.

## Phase Transition Analysis on DFAs

In this section, we focus on the structure of DFA to explain why the easy-hard-easy pattern occurs. DFA can also be used as a target language. In the early study, DFA was used to solve CSPs by Vempary [24]. Amilhastre et al. [17] compiled configuration problems into DFAs. Such a DFA is a directed graph, its edges (the transitions of the automaton) are labeled by an assignment <$x_i,a_i$>, which means $a_i$ is assigned to $x_i$. Nodes of DFAs are called states; the root of the graph is called the initial state. States in the graph are divided into subsets (levels), where the initial state composes level 0; labels of transitions coming from states in the same subset have the same variable but may have different values. Paths from the initial state to a final state are of the same length. Such a path represents a solution of the SAT (or CSP) instance. It is known that the DFA is a special form of the BDD, but the structure of DFAs is more regular than that of OBDDs. Because of the regular structure, we choose it to analyze the reason why the easy-hard-easy pattern occurs by studying its microstructure.

Each clause in a SAT instance is a disjunction of literals, where a literal is either a variable or the negation of a variable. We often denote a clause with $k$ length by $x_{m_1} \vee \ldots \vee x_{m_j} \vee \neg x_{m_{j+1}} \vee \ldots \vee \neg x_{m_k}$ in this section. The *adjoint nogood* of the clause is the partial assignment $(<x_{m_1},0>,\ldots,<x_{m_j},0>,<x_{m_{j+1}},1>,\ldots,<x_{m_k},1>)$. A path is compatible with a partial assignment if all pairs of <$x_i,a_i$> in the partial assignment occur in some labels of the path. Adding a clause to a DFA means that paths compatible with its adjoint nogood are removed. An edge $e$ is an *adjoint edge* of a path if the edge is not in the path, and there exists an edge $e'$ in the path such that $e'$ and $e$ come from a common state and point to a common state. A path is an *i-interchangeable path* of a clause $l$ if there exist only $i$ adjoint edges whose variables (all variables are different) contained in the adjoint nogood of the clause and the path is compatible with the nogood. A path without such an adjoint edge is regard as a 0-interchangeable path of the clause. A 1-interchangeable path is called a *single interchangeable path*, whereas *multi-interchangeable paths* are $i$-interchangeable paths such that $i$>1.

### Phase transition on interchangeable path of DFAs

When a clause is conjoined to a DFA, paths compatible with the adjoint nogood of the clause have to be removed from the DFA. Consider a single path in a DFA that will be removed, 3 cases have to be investigated. If the path is a 0-interchangeable path of the clause, the size of the DFA probably decreases, because some nodes in the path should be deleted and no extra nodes are added. Fig. 6(a) shows an example of this procedure. If the path is a single-interchangeable path, the size of the DFA changes very slightly, because the nodes of the path always remain (Fig. 6(b)). If the path is a multi-interchangeable path, the size of the DFA usually increases, because the path has more than one adjoint edge. To remove the path, other paths containing adjoint edges should be separated from the removed path so that extra nodes are required. Fig. 6(c) shows an example of this procedure.

We only show some simple cases, while there may be more than one path compatible with adjoint nogood of a clause. Besides, it is known that dynamic variable ordering on BDDs or DFAs has a great impact on the sizes, but analyzing on the DFA sizes under those situations is extremely complex, so we only consider basic cases in the microstructure analysis.

Because of the structural complexity, we first require some approximate suppositions. We suppose that the size will increase when adding a clause that has one or more multi-interchangeable paths, and a DFA is in the easy-hard phase if more than half of clauses make the size of the DFA increase, otherwise it is in the hard-easy phase. Next, the following experiment is performed. We generate SAT instances with $k=3$ and $n=25$. For each $r$ from 0 to its $r_p$ (4.3), 100 instances are generated. For each instance, we construct the DFA, and then generate 100 clauses randomly and uniformly. The number of clauses that have multi-interchangeable paths is counted. If the number is larger than 50, we regard that the instance is in the easy-hard phase. Fig. 7(a) shows the results, where the curve with circle nodes depicts the number of instances in the easy-hard phase for each $r$, and the curve without nodes shows the average number of DFA nodes. Similar experiments are also performed with parameter $k=4$ and $n=18$, $k=5$ and $n=15$, $k=7$ and $n=12$ respectively. Fig. 7(b) (c) and (d) depict the results.

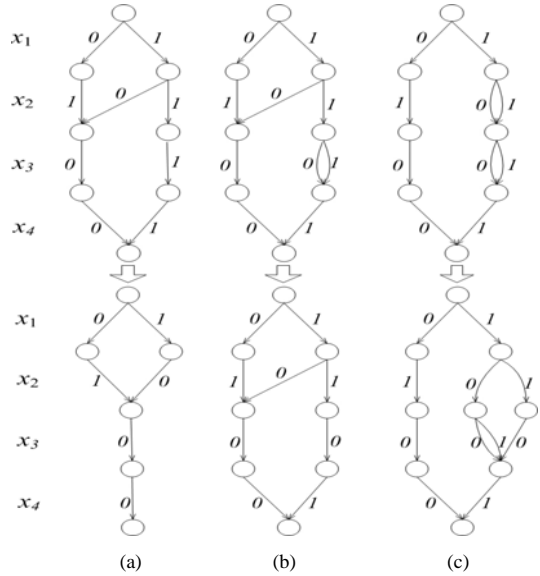

**Fig. 6.** Cases of conjoining clauses to DFAs. DFAs in the first row are original ones, and DFAs in the second row are resultant ones after adding clause $\neg x_1 \vee \neg x_2 \vee \neg x_3$.

From those figures, we can observe that the number of instances have a sharp transition from the increment phase to the decrement phase. It is also worth to mention that the areas of the sharp transition are very close to the peak points of the easy-hard-easy pattern, though they are not very exact because of our approximate supposition. Therefore, we can conclude that an important factor that makes the DFA sizes suffer an easy-hard-easy pattern is the existence of multi-interchangeable paths. Because of separations on the multi-interchangeable paths when adding clauses to a DFA, more states have to be constructed for the resultant DFAs. Besides, it is noted that the peak points are compatible with the results of the OBDD and d-DNNF.

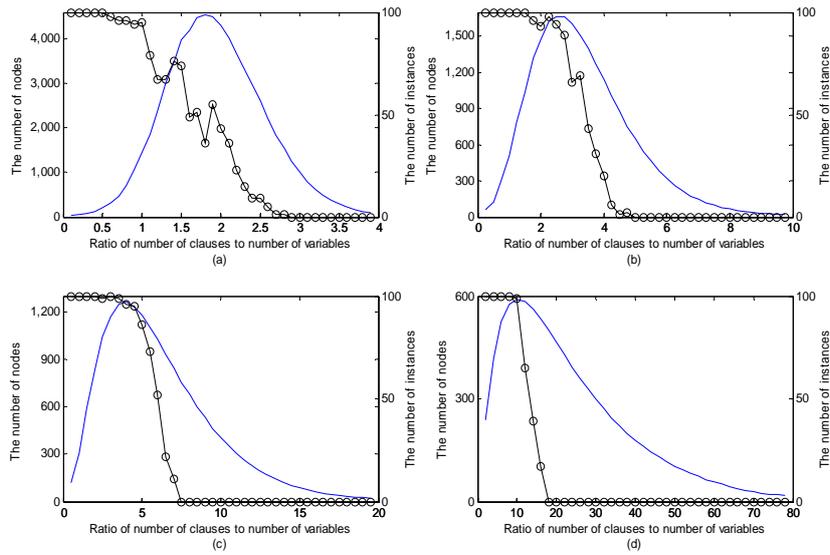

**Fig. 7.** Phase transitions of multi-interchangeable paths in *k*-SAT.

**Solution Interchangeability**

We then discuss the relationships between the multi-interchangeable paths and the solution interchangeability. Solution interchangeability of CSPs has been studied for decades. Freuder [25] investigated the interchangeability deeply, where various forms of interchangeability were proposed. According to his definitions, two values *a, b* for a variable *x* are *fully interchangeable* if every solution to the CSP containing the assignment <*x, a*> remains a solution when *b* is substituted for *a*, and vice versa. Here we extend the solution interchangeability by the following definitions: a SAT instance with *n* variables is *fully multi-interchangeable* if there exists more than one variable such that values 0 and 1 are interchangeable for every solution; a partial assignment is *weak interchangeable* if it can be extended to a solution by assigning any value (0 or 1) to each unsigned variable; a partial assignment is *weak multi-interchangeable* if it is *weak interchangeable* and its length is no more than *n*-2 (*n* is the total number of variables in an instance). Clearly, a multi-interchangeable path implies that there must exist at least one multi-interchangeable partial assignment, but a multi-interchangeable partial assignment does not mean there exists a multi-interchangeable path. That is, the definition of the multi-interchangeable path is more rigorous than that of the weak multi-interchangeable partial assignment. On the other hand, it can be seen that the definition of the multi-interchangeable path is looser than that of fully multi-interchangeable SAT instance. So the multi-interchangeable path is a more complex form of solution interchangeability between full and weak solution interchangeability.

We can also see that the solution interchangeability in SAT instances is related to implicants [26]. A term $\kappa$ is an implicant of the formula $\varphi$ if and only if $\kappa \models \varphi$. Implicants are a dual form of the implicates. The interchangeable paths mean that there exist some implicants on the path, while an implicant may not appear in the form of an interchangeable path. In fact, when we use OBDD, d-DNNF and DFA as the target languages, the key technique during the compilation is to convert solutions into a compact form. Because of the compaction, compilation results can be reduced to reasonable sizes for on-line processing. The compaction can reduce the representation of symmetric solutions by incorporating some common parts of these solutions. So the interchangeable structure of solutions determines the effectiveness of the compaction, and thus the transition of the solution interchangeability mentioned above causes the sizes of compilation results suffering the easy-hard-easy pattern.

## Conclusions

Our work concerns the phase transition in knowledge compilation of *k*-SAT, which is accompanied with an easy-hard-easy pattern on the space rather than the runtimes. We first conjectured that the ratios of clauses to variables in the random SAT model have great impact on the sizes of compilation results, where an easy-hard-easy pattern exists similar with many phase transition phenomena. DFAs, OBDDs and d-DNNFs, which are commonly employed to compile real-world problems off-line, have been chosen as the target languages for compilation. Experimental results on randomly generated *k*-SAT have been demonstrated in a number of figures, by which the conjecture has been supported. Interestingly, we suggested curves of average sizes using different target languages have peak points with the same value of *r* when *k* is fixed. It can also be concluded that the sizes for 3-SAT increase exponentially as *n* grows when *r*>0.3, whereas the sizes increase polynomially when *r*<0.3. Hence, there exists another phase transition that separates the regions of polynomial and exponential sizes. Furthermore, we try to explain the reason that the existence of the easy-hard-easy pattern. From approximate analyses on the structures of DFAs, we concluded that paths with more than one adjoint edge have great impact on the sizes of DFAs, and these paths imply a kind of solution interchangeability in the SAT instances. As can be seen that those paths has strong relation with implicants, further works will include uncovering more inherent relations between them.

**Acknowledgements.** We would like to thank Nina Narodytska for her BDD compiler. The work described in this paper was supported by the National Natural Science Foundation of China Granted No. 60803102 and 60973033.